\documentclass[runningheads]{llncs}
\usepackage{adjustbox}
\usepackage{amsmath}
\usepackage{amssymb}
\usepackage{float}
\usepackage[T1]{fontenc}
\usepackage{graphicx}
\usepackage{hhline}
\usepackage[utf8]{inputenc}
\usepackage{mathtools}
\usepackage{multicol}
\usepackage{multirow}
\usepackage[normalem]{ulem}
\usepackage{wasysym}
\usepackage[hidelinks]{hyperref}
\usepackage{xurl}

\usepackage{xcolor}

\begin{document}
\title{Automatic Authorship Attribution in the work of Tirso de Molina
}
%
%
\author{Miguel Cavadas Docampo\inst{1}\and Pablo Gamallo\inst{2}}
\authorrunning{M. Cavadas and P. Gamallo}
%
\institute{Universidade de Santiago de Compostela, \email{miguel.cavadas@rai.usc.es}
\and Centro de Investigaci\'on en Tecnolox\'ias da Informaci\'on (CiTIUS)\\
  Universidade de Santiago de Compostela,  \email{pablo.gamallo@usc.gal}}
\maketitle              

\begin{abstract}
Automatic Authorship Attribution (AAA) is the result of applying tools and techniques from Digital Humanities to authorship attribution studies. Through a quantitative and statistical approach this discipline can draw further conclusions about renowned authorship issues which traditional critics have been dealing with for centuries, opening a new door to style comparison. The aim of this paper is to prove the potential of these tools and techniques by testing the authorship of five comedies traditionally attributed to Spanish playwright Tirso de Molina (1579$-$1648): \textit{La ninfa del cielo}, \textit{El burlador de Sevilla}, \textit{Tan largo me lo fiáis}, \textit{La mujer por fuerza} and \textit{El condenado por desconfiado}. To accomplish this purpose some experiments concerning clustering analysis by Stylo package from R and four distance measures are carried out on a corpus built with plays by Tirso, Andrés de Claramonte (c. 1560$-$1626), Antonio Mira de Amescua (1577$-$1644) and Luis Vélez de Guevara (1579$-$1644). The results obtained point to the denial of all the attributions to Tirso except for the case of \textit{La mujer por fuerza}.
\end{abstract}

\section{Introduction}\label{sec:intro}

Authoring problems have been a constant throughout the history of literature. Examples of doubtful authorship constitute an important percentage of the canonical works of world literature as a whole, whether by legal prohibitions, improper appropriation, or collaborations between several authors. The Digital Humanities, absorbing methods from natural language processing and theories from literary research, have emerged as an useful framework to deal with problems of authorship, giving rise to the discipline known as \textit{non$-$traditional authorship attribution studies} or also \textit{automatic authorship attribution} (hereinafter referred to as AAA). This discipline, far from disavowing traditional studies, serves as a complement to these, identifying quantifiable markers of authorship (Almeida, 2014, p. 10). AAA is based on the following assumption: $``$the author of a text can be selected from a set of possible authors by comparing the values of textual measurements in that text to their corresponding values in each author’s writing sample$"$ (Grieve, 2005, p. 1). The growing interest in AAA has given rise to an international workshop organized as a set of related shared tasks and held since 2011 as part of the PAN Lab on Uncovering Plagiarism, Authorship, and Social Software Misuse (Argamon $\&$ Juola, 2011). The main objective of PAN shared tasks is to propose a standardized evaluation framework for authorship attribution, by means of which is possible to compare the strategies that participate in the event. 

The objective of this article is, first of all, to highlight the role that certain tools coming from computational linguistics may have in an authorship study and, consequently, to point out the value of AAA as a profitable convergence between disciplines. In order to illustrate this with an example, we will focus specifically on the Tirsian debate with the aim to draw conclusions that may serve as relevant arguments in order to reinforce some of the most supported critical positions. More in particular, we aim to find the most probable authors of five plays from the Golden Age that were traditionally attributed to Tirso de Molina, among them the famous piece entitled \textit{El burlador de Sevilla y el convidado de piedra} (\textit{The Trickster of Seville and the Stone Guest}), which is the source of the myth of the iconic lover Don Juan. All these five plays have been surrounded, especially in the last decades, by great controversy regarding their alleged authorship. To perform such experiments, different AAA strategies were configured and applied on a corpus consisting of twelve plays attributed to four different authors, including Tirso de Molina, and five plays about which there are doubts concerning authorship.

The main contribution of our work is, in general terms, the redirection of the Tirsian debate towards the AAA sphere and the strengthening of the critical positions which deny Tirso’s authorship in several plays traditionally attributed to him, with the consequent restitution to the original authors. We are forced by the results to highlight the figure of Andrés de Claramonte, who we state as most likely author of Tirso’s famous play \textit{El burlador de Sevilla}. This implies two debates: one about the position of Tirso de Molina in the canon of Spanish Golden Age literature and another about how authorship studies were performed in the past. Taking into account the method we follow, the application of non$-$stylometric computational strategies to the field of authorship attribution is also a remarkable contribution since really promising results are obtained.

The article is organized as follows: Section 2 offers a brief comparison between the way authorship attribution studies were performed in the past and the new techniques digital humanities are tuning up. Section 3 includes a thorough description of the means and methods we followed in order to design the experiments with the Tirsian corpus, which we explain, execute and discuss in Section 4. Thus relevant facts about authorship in Tirso de Molina’s theatre are given in Section 5, as well as the guidelines that further investigation should follow in order to face this fruitful and exciting debate.

\section{Background and related work}

Researchers who have ventured into the rough terrain of authorship attribution studies have developed and refined a great number of methods based on statistical comparison of various style markers that allegedly shed light on each writer's idiolect. These markers include metrics, word or sentence length, punctuation, lexical indexes, errors, word classes, collocations, and other morphosyntactic patterns. However, analysis of these elements has generally not led to any firm conclusions, revealing their methodological weakness. The main obstacle to resolution may be that, in many cases, the method used does not have a sufficiently decisive degree of differentiation to make a solid proposal of attribution. It is precisely this point where AAA may come in and achieve greater power due to the mathematization of texts. The main difference AAA intends to make from conventional studies is that $``$only quantitative techniques may be empirically evaluated and mechanically applied$"$ (Grieve, 2005), so its aim is not to replace traditional studies but to complement them with numeric methodologies.

AAA is a discipline that has been under development for about fifty years. Even so, it has not been free of criticism, controversy and continuing problems, among which Rudman cites: $``$studies governed by expediency; a lack of competent research; flawed statistical techniques; corrupted primary data; lack of expertise in allied fields; a dilettantish approach; inadequate treatment of errors$"$ (Rudman, 1998, p. 351). These weaknesses are generally due to the scientific perspective with which the NTAAS need to be approached, as opposed to other humanistic research. The following statement by Rudman makes it clear: $``$Every non$-$traditional authorship attribution study is an experiment, a 'scientific' experiment$"$ (Rudman, 2016, p. 310), so it must strictly comply with two principles: replicability and an experimental plan. The solutions Rudman proposes to this deficiency are: $``$construct a correct and complete experimental design; educate the practitioners; study style in its totality; identify and educate the gatekeepers; develop a complete theoretical framework; form an association of practitioners$"$ (Rudman, 1998, p. 351).

To date, most AAA research has been conducted only in English, with multitude of studies that aim to define the authorship of Shakespearean texts (some of them reviewed by Brian Vickers in 2011). Very few papers focus on old Spanish literature because of the lack of literary texts available in acceptable formats, which forces each researcher to take charge of obtaining a dignified text from different HTML pages (Calvo Tello, 2016). In spite of these obstacles, the two most famous author dilemmas in the history of Spanish literature have already been tackled under this approach: Javier de la Rosa and Juan Luis Suárez (2016) obtained results that point to the jurist Juan Arce de Otálora as probable author for \textit{Lazarillo}, while Javier Blasco (2016) associated the \textit{Apocryphal Quixote}, in an attempt to unmask Avellaneda, with writers such as Castillo Solórzano or Tirso de Molina. More recently, in 2017, the project EstilometríaTSO was created in order to work on stylometric aspects of Spanish Golden Age theatre (Cuéllar González $\&$ García$-$Luengos, 2017). Given the little work done so far in this field, it seems clear that this research is still in a very incipient phase.

The theatre texts of the Spanish Golden Age, due to the intricate process of textual transmission that most of them suffered, exhibit an extra difficulty in the process of authorship attribution. The production of Tirso de Molina (1579$-$1648), a pseudonym by which Fray Gabriel Téllez is known, is a clear example of these open$-$ended dilemmas. Many critics have devoted their studies to clarifying the various problems of authorship on Tirso's work, but in general the results are by no means conclusive. 

Two AAA methods have been defined: profile and instance$-$based approaches. The profile$-$based approach does not consider the differences between texts written by the same author, as it uses for each author a single huge file built by chaining all the texts available to that author. The instance$-$based approach requires multiple training text samples per author in order to develop an accurate attribution model (Stamatos, 2009). About which of the two approaches is the most effective, Nektaria Potha and Efstathios Stamatos, paying attention to the results of PAN$-$2013 authorship verification competition, conclude that instance$-$based approaches are more appropriate for this task. However, they admit that $``$the profile$-$based paradigm is more effective when only short and limited samples of documents are available$"$ (Potha $\&$ Stamatos, 2014, p. 314). We follow both the profile$-$based approach, since our corpus is limited, and the instance$-$based one, because of its better achievements. 

The stylistic features AAA uses rely mainly on the frequency of usage of function words. Eisen et al. have developed a new technique based on function word adjacency networks (WANs) $``$with function words as nodes, and edges containing information regarding the use of two function words within a certain distance$"$ (Eisen et al., 2018). To examine the use of such WANs R Stylo features Rolling Delta and Rolling Classify may be used, as does Hartmut Ilsemann in both his recent studies about the authorship of the \textit{Parnassus Plays }(Ilsemann, 2018) and Thomas Kyd’s \textit{Cornelia }(Ilsemann, 2019). In fact, Ilsemann makes use of a powerful command of the software, taking advantage of all its features, including multidimensional scaling, bootstrap consensus tree and cluster analysis (the one we draw on in this paper). He warns that $``$although the various features of the R Stylo application offer a huge number of results according to the number of chosen parameters, the overall assessment can be seen as an external proof of authorship$"$ (Ilsemann, 2018, p. 555), hence his encouragement of the rollout of those features in order to build a \textit{forensic }(etymologically) discipline, that is, open to public debate.

\section{The method}

AAA relies on different statistical measures that take into account the distribution of function words. Function words not only act as a lexical indicator, but also contain a high degree of syntactic information, as they participate in many syntactic structures. Extracting the most frequent words from a text will be the first step to perform a stylometric comparison between documents. The most important criterion for selecting features in authorship attribution tasks is their frequency. In general, the more frequent a feature, the more stylistic variation it captures (Stamatos, 2009). The distribution of the most frequent terms (function words) in the texts constitutes the author's footprint. The footprint of an author is not the specific words he/she uses the most, but how much he/she uses them. The main differences between AAA strategies lie in the selection of the appropriate measure to compare the documents on the basis of the word distribution (or author's footprint).  The selection of the most appropriate measure depends on the available corpus and the objective of the study. However, according to Grieve (2005), the best approach to quantitative authorship attribution is one that is based on the values of as many textual measurements as possible. This is because, in general, small variations in the configurations can produce very large changes in the results. For this reason, five different strategies are used in this work. The five strategies are divided in two opposing approaches: instance and profile$-$based.

\subsection{The instance$-$based approach}

Individual texts are the basic items of the procedure. Following this approach, we use clustering analysis on individual texts so as to group them on the basis of stylistic similarities, which is equivalent, if the adjustments are correct, to group the texts by author. The resulting grouping of individual texts is displayed in a dendogram. The clustering process is based on the Delta measure, created by John Burrows in 2002 specifically for stylometric purposes. This method was implemented by Stylo, a flexible R package for the high$-$level stylistic analysis of text collections (Eder, 2016).  The Delta measure normalizes frequencies by means of z$-$score to reduce the influence of very frequent words. For f\textsubscript{i}(D) being the frequency of n$-$gram\textsubscript{i} in document D, $\mu$\textsubscript{i} the mean frequency of the n$-$gram in the corpus, and $\sigma$\textsubscript{i} its standard deviation, then z$-$score is defined as follows:

\begin{equation}
z\left(f_{i}\left(D\right)\right) =\frac{\left(f_{i}\left(D\right) - \mu_{i}\right)}{\sigma_{i}} 
\end{equation}

The difference between a set of training texts written by the same author and an unknown text is the mean of the absolute differences between the \textit{z}$-$scores (Stamatos, 2009)Given the normalized document vectors, the Burrows's Delta is just the Manhattan distance by using normalized frequencies with z$-$scores. Given documents D\textsubscript{1} and D\textsubscript{2} , distance Delta $\Delta$ is computed as follows:

\begin{equation}
\Delta  = \sum_{\text{i = }1}^{n}\left\vert z\left(f_{i}\left(D_{1}\right)\right) - z\left(f_{i}\left(D_{2}\right)\right)\right\vert
\end{equation}

The lower the Delta value the higher the similarity between the texts studied. Recently, Maciej Eder (2016) has made some innovations on the classical method of Burrows. In his experiments, José Calvo Tello (2016) proves that these changes yield better results. Stylo allows the user to choose both Burrows' and Eder's measures and even others based on different similarity methods, such as Wuzburg distance. Since they all provide analogous results, as we have tested, we will limit to Burrows and Eder.

\subsection{The profile$-$based approach}

In this approach, the known texts belonging to one author are merged into one single document (profile of the author) and, then, a distance measure is computed between the profile of the author and the profile of an unknown text. Four different distance measures were designed and implemented: Kullback$-$Leibler divergence, Perplexity, Ranking$-$based distance, and Distributional similarity. These measures represent four different corpus$-$based strategies to compare texts. They were not originally designed to serve the purposes of AAA, but are commonly employed in other computational tasks such as language identification, language distance, information retrieval and data mining.

\subsubsection{\textit{Kullback$-$Leibler}}

Kullback$-$Leibler divergence compares two distributions, more precisely, is a measure of how one probability distribution (for instance, the profile of an unseen document) is different from a second, reference probability distribution (the profile of the author). In Iriarte et al. (2018) it was used to measure the distance between texts written by different social groups of individuals: men / women, university / non$-$university people, and so on.  Given a test or unknown text (T) and the known texts of an author (A), the Kullback–Leibler divergence KL of the distributions T and A  is defined as follows:

\begin{equation}
KL\left(A,T\right) = \sum_{}^{}A\left(ngr_{i}\right)\log \frac{A\left(ngr_{i}\right)}{T\left(ngr_{i}\right)} 
\end{equation}

\vspace{1\baselineskip}
Equation 3 allows computing how far the T distribution is from the A distribution, taking into account the probabilities of the n$-$grams (of words or characters) in each compared text corpus, either T or A. The divergence (which is an asymmetric measure) was converted into a symmetric one (i.e. into a distance) by computing the mean of the two complementary comparisons: divergence of X from Y, and divergence of Y from X. In our experiment, we applied Kullback$-$Leibler divergence on distributions of the most frequent word unigrams. \ \

\subsubsection{\textit{Perplexity}}

Perplexity is frequently used as a quality measure for language models built with n$-$grams extracted from text corpora, and can be used to measure how well a model (for instance, the profile of an author) fits the test data (the profile of an unseen document). More formally, perplexity is the normalized inverse probability of an input test. It can be used to compare a test text (T) with the author model (M). The perplexity PP of T given the author model M is defined by the following equation:

\begin{equation}
PP\left(M,T\right) = 2^{ - \sum_{}^{}T\left(ngr_{i}\right)\log _{2}M\left(ngr_{i}\right)}
\end{equation}
where ngr\textsubscript{i} is a n$-$gram shared by both T and M. Equation 4 can be used to set the divergence between a test set and the author model. The lower is the perplexity of T given M, the lower is the distance between the two compared objects. Texts may be modeled with n$-$grams of either words or characters. In our experiments, we applied PP distance to texts modeled with 7$-$grams of characters. In other pieces of work, PP was also used to compare the linguistic distance between 40 European languages (Gamallo et al. 2017), as well as to compute the distance between diachronic varieties of the same language (Pichel et al. 2019).

\subsubsection{\textit{Rank$-$Based}}

The rank$-$based distance between two languages is obtained by comparing the ranked lists of the two languages. It takes two word profiles (the author and the unseen document) and calculates a simple rank$-$order statistic based on an $``$out$-$of$-$place$"$ measure. This measure determines how far out of place an n$-$gram in one profile is from its place in the other profile (Cavnar $\&$ Trenkle, 1994). This measure is often used to compute language identification (Gamallo et al., 2014). More formally, given the ranked lists Rank\textsubscript{T} and Rank\textsubscript{A} of the test text (T) and the texts of a given author (A), respectively, the rank$-$based distance, R, is computed as follows:

\begin{equation}
 R\left(A,T\right) = \sum_{i = 1}^{K}\left\vert Rank_{A}\left(ngr_{i}\right) - Rank_{T}\left(ngram_{i}\right)\right\vert
\end{equation}
where K stands for the number of the most frequent n$-$grams , Rank\textsubscript{A}(ngr\textsubscript{i}) is the rank of a specific n$-$gram, ngr\textsubscript{i}, in A, and Rank\textsubscript{T}(ngr\textsubscript{i})  is the rank of the same n$-$gram in T. In our experiments, we applied this distance on lists of word unigrams.

\subsubsection{\textit{Cosine similarity}}
As a forth measure, we use vector space models, which is one of the most popular representation of document vocabulary, to compute distributional similarity between documents. In particular, this strategy compares the word vectors extracted from the author's profile with those extracted from the unseen document. Given that it is a measure of similarity, the higher its value, the higher will be the similarity between the texts, being the maximum value 1. As it works in an inverse way to the other three, which are distance measures, we turn it into a distance by subtracting the values from 1. In our experiments, we used Cosine similarity by considering the comparative study made by Afzali $\&$ Kumar (2017) on different metrics, namely Cosine, Jaccard and Dice, to evaluate their performance in finding the similarity of two text documents. Cosine outperformed the other metrics in a significant way. Given A and T, Cosine (as a distance) is defined in this way:

\begin{equation}
Cosine\left(A,T\right) = 1 -\frac{\sum_{}^{}A\left(ngr_{i}\right)T\left(ngr_{i}\right)}{\sqrt{\sum_{}^{}A\left(ngr_{i}\right)^{2}}\sqrt{\sum_{}^{}T\left(ngr_{i}\right)^{2}}}
\end{equation}

\subsection{\textit{Mean of measures}}
Finally, the four measures defined above are merged by averaging their values. In order to compute the mean of the values obtained by the four distance measures, their final scores are normalized. The four measures and its mean combination have been implemented by the authors into an executable written in PERL language. The software is aimed at perform authorship attribution and is freely available.\footnote{\url{https://github.com/gamallo/Autoria}}

It must be stressed that the use of four different profile$-$based strategies, in addition to the instance$-$based one defined above, allows us to reach more solid and reliable conclusions about authorship. These are not different configurations of a single strategy, as is usually done in most Stylo$-$based work, but five complementary methods of covering diverse aspects of the same problem.

\section{The experiments}\label{sec:experiments}
\subsection{Tirso de Molina and authorship issues on part of his work}
Edition and diffusion of the work of Tirso is a clear example of the complex process underlying the transmission of Golden Age plays. Comedies were printed either in blocks called \textit{partes} (generally twelve plays each), or in individual pamphlets, after having been represented and passing through the hands of at least the director and the publisher, without the poet having any control at any stage of the procedure. It is not surprising, therefore, that authorship misunderstandings arose during the process. Of the five blocks that were published under the name of Tirso, critics warn of the unreliability of two of them, the second and third ones, published in the 1630s. In the case of Block Two, which contains twelve plays, Tirso himself warns of the fraudulent editorial maneuver: $``$Dedico, destas doce comedias, cuatro que son mías en mi nombre, y en el de los dueños de las otras ocho (que no sé por qué misfortunio suyo, siendo hijas de tan ilustres padres las echaron a mis puertas), las que quedan$"$ (Oteiza, 2000, p. 106).\footnote{$``$I dedicate, from these twelve plays, from which only four of them are mine under my name, to the owners of the other eight (unaware of the misfortune that brought them, being daughters of such distinguished fathers, in front of my doors) the remaining ones$"$.} Unfortunately, Tirso does not report the actual authorship of the eight works that he did not write. To this day, critics have been able to elucidate the authorship of several plays. The current situation about authorship attribution of Block Two is depicted in Table 1. 

\begin{table}[h]
\begin{adjustbox}{max width=\textwidth}
\begin{tabular}{p{7.62cm}p{7.73cm}p{0.07cm}}
\hline
\multicolumn{1}{|p{7.62cm}}{\centering
\textbf{Comedia}} & 
\multicolumn{1}{|p{7.73cm}|}{\centering
\textbf{Autor}} \\ 
\hline
\multicolumn{1}{|p{7.62cm}}{\textbf{\textit{La reina de los reyes}}} & 
\multicolumn{1}{|p{7.8cm}|}{Hipólito de Vergara} \\ 
\hline
\multicolumn{1}{|p{7.62cm}}{\textbf{\textit{Amor y celos hacen discretos}}} & 
\multicolumn{1}{|p{7.73cm}|}{Tirso de Molina} \\ 
\hline
\multicolumn{1}{|p{7.62cm}}{\textbf{\textit{Quién habló pagó}}} & 
\multicolumn{1}{|p{7.8cm}|}{Rodrigo de Herrera} \\ 
\hline
\multicolumn{1}{|p{7.62cm}}{\multirow{2}{*}{\parbox{7.62cm}{\textbf{\textit{Siempre ayuda la verdad}}}}} & 
\multicolumn{1}{|p{7.73cm}|}{Before: Luis Belmonte Bermúdez? Juan Ruiz de Alarcón? Rodrigo de Herrera? Tirso de Molina?} \\ 
\hhline{~-}
\multicolumn{1}{|p{7.62cm}}{} & 
\multicolumn{1}{|p{7.8cm}|}{Since García$-$Reidy (2019): Lope de Vega} \\ 
\hline
\multicolumn{1}{|p{7.62cm}}{\textbf{\textit{Los amantes de Teruel}}} & 
\multicolumn{1}{|p{7.73cm}|}{Juan Pérez de Montalbán} \\ 
\hline
\multicolumn{1}{|p{7.62cm}}{\textbf{\textit{Por el sótano y el torno}}} & 
\multicolumn{1}{|p{7.8cm}|}{Tirso de Molina} \\ 
\hline
\multicolumn{1}{|p{7.62cm}}{\textbf{\textit{Cautela contra cautela}}} & 
\multicolumn{1}{|p{7.73cm}|}{Mira de Amescua} \\ 
\hline
\multicolumn{1}{|p{7.62cm}}{\textbf{\textit{La mujer por fuerza}}} & 
\multicolumn{1}{|p{7.8cm}|}{Tirso de Molina?} \\ 
\hline
\multicolumn{1}{|p{7.62cm}}{\textbf{\textit{El condenado por desconfiado}}} & 
\multicolumn{1}{|p{7.73cm}|}{Mira de Amescua? Andrés de Claramonte? Vélez de Guevara? Tirso de Molina? Collaboration?} \\ 
\hline
\multicolumn{1}{|p{7.62cm}}{\textbf{\textit{Primera parte de don Álvaro de Luna}}} & 
\multicolumn{1}{|p{7.8cm}|}{Mira de Amescua} \\ 
\hline
\multicolumn{1}{|p{7.62cm}}{\textbf{\textit{Segunda parte de don Álvaro de Luna}}} & 
\multicolumn{1}{|p{7.73cm}|}{Mira de Amescua} \\ 
\hline
\multicolumn{1}{|p{7.62cm}}{\textbf{\textit{Esto sí que es negociar}}} & 
\multicolumn{1}{|p{7.8cm}|}{Tirso de Molina} \\ 
\hline
\end{tabular}
\end{adjustbox}\\
\caption{Comedies in Tirso’s \textit{Parte Segunda} with their corresponding authors or most likely candidates (before question mark)}
\end{table}

The main doubts about authorship lie on two plays: \textit{La mujer por la fuerza} (\textit{The woman by force}) and \textit{El condenado por desconfiado }(\textit{The condemned by distrust}). A multitude of arguments have been proposed for the second one, rejecting Tirso's authorship and assigning it to other authors. The disagreement between critics is extreme and even the hypothesis of collaboration is considered (Rodríguez López$-$Vázquez, 2010). It should be noted that recently, Alejandro García$-$Reidy (García$-$Reidy, 2019) seems to have demonstrated Lope's authorship of the work \textit{Siempre ayuda la verdad }(\textit{Always helps the truth}), traditionally attributed to Tirso, using, among other means, the Stylo software by R, which has also been used to carry out some of our experiments.

The most important authorship problems concern single editions of pamphlets, que fue el medio de transmission de \textit{La ninfa del cielo}, which has been returned to Luis Vélez de Guevara in the critical edition of Alfredo Rodríguez López$-$Vázquez (2008) for Cátedra without any refutation so far and the well-known \textit{El burlador de Sevilla} (\textit{The Burlador of Seville}) and its alternate version named \textit{Tan largo me lo fiáis}, coinciding in 1433 verses, which represent more than 60$\%$ of the total (García Gómez, 2005). It is important to point out that Rodríguez López$-$Vázquez (Rodríguez López$-$Vázquez, 1990) attributes the authorship of \textit{Tan largo} (and consequently that of \textit{El burlador}) to playwright Andrés de Claramonte (c. 1560$-$1626), who stands as a central figure in this debate, despite the fact that his name may not be familiar at all. According to Rodríguez López$-$Vázquez, the cause of the scarcity of bibliography on this author, his absence from the histories of literature and the large portion of his production that remains unedited must be traced by the fact that he has been seen as a plagiarizer. In this way, the resolution of the problem of authorship of \textit{El Burlador} becomes essential for determining the place that Claramonte should occupy in the history of literature.

\subsection{The corpus}

The ideal corpus to which these methods should be applied would include all the works of Tirso whose authorship is in dispute and several representative works of all possible authors. However, the difficult availability of digitalized works limits greatly the creation of the corpus as there is a very large lack of digitalizations of even the most recognized authors. The fraction of Golden Age comedies that can be freely consulted online is very small. The source in which most of them are stored is the web site of the Miguel de Cervantes Virtual Library (BVC),\footnote{\url{http://www.cervantesvirtual.com/}} which offers a multitude of literary and critical texts of the entire history of Spanish literature. The website of the Association for Hispanic Classical Theater (AHCT)\footnote{\url{http://www.wordpress.comedias.org/}} is more focused on Golden Age comedies and, then, contains a significant variety of authors of this period. It should be noted that the disadvantage of the lack of digitalization is the direct consequence of another much more serious problem: the shortage of editions. This phenomenon of scarcity is especially serious in the case of Andrés de Claramonte, to whom critics have paid little attention, and makes it enormously difficult to set up an extensive corpus to study attributions.

For all this, the main criterion used for building a corpus that serves to examine the problems of authorship of the theatre of Tirso de Molina is necessarily the availability of the texts. Table 2 shows the list of plays with dubious authorship that we were able to convert into text files ready to be computationally processed.  

\begin{table}[h]
\begin{adjustbox}{max width=\textwidth}
\begin{tabular}{p{7.62cm}p{7.71cm}}
\hline
\multicolumn{1}{|p{7.62cm}}{\centering
\textbf{Text}} & 
\multicolumn{1}{|p{7.71cm}|}{\centering
\textbf{Possible authors}} \\ 
\hline
\multicolumn{1}{|p{7.62cm}}{\textbf{\textit{El burlador de Sevilla}}} & 
\multicolumn{1}{|p{7.71cm}|}{Claramonte / Tirso} \\ 
\hline
\multicolumn{1}{|p{7.62cm}}{\textbf{\textit{El condenado por desconfiado}}} & 
\multicolumn{1}{|p{7.71cm}|}{Amescua / Claramonte / Guevara / Mira / Tirso / Colaboration} \\ 
\hline
\multicolumn{1}{|p{7.62cm}}{\textbf{\textit{La mujer por fuerza}}} & 
\multicolumn{1}{|p{7.71cm}|}{Tirso / Other (no name proposed)} \\ 
\hline
\multicolumn{1}{|p{7.62cm}}{\textbf{\textit{La ninfa del cielo}}} & 
\multicolumn{1}{|p{7.71cm}|}{Guevara / Tirso} \\ 
\hline
\multicolumn{1}{|p{7.62cm}}{\textbf{\textit{Tan largo me lo fiáis}}} & 
\multicolumn{1}{|p{7.71cm}|}{Claramonte / Tirso} \\ 
\hline
\end{tabular}\\
\end{adjustbox}\\
\caption{Texts under discussion and their corresponding possible authors.}
\end{table}

As can be deduced from Table 2, the authors involved in the study are, apart from Tirso, Andrés de Claramonte, Mira de Amescua and Luis Vélez de Guevara. 

The comedies of the Golden Age are texts quite similar to each other because they respond to the same stylistic and narrative schemes. In these conditions, given that discrimination between authors is more complicated, it is necessary to compile texts that are very representative of each author's style. In order to accomplish this, we have sticked to the traditional appraisal of the texts, picking the most familiar titles among the available ones. In those cases where there were sufficient texts (Tirso and Mira), we selected plays whose date of composition was as close as possible to that of works of dubious authorship, trying to respect the principle of $``$all the texts per author should be written in the same period to avoid style changes over time$"$ (Stamatos, 2009, p. 558). The BVC has been used as the main source of texts, so that almost all the texts have been edited based on similar criteria. When the comedy was not available on that web site, we have resorted to the AHCT or, failing that, to editions. Finally, the conformed corpus is the one detailed in Table 3.

\begin{table}[h]
\begin{adjustbox}{max width=\textwidth}
\begin{tabular}{p{3.08cm}p{5.13cm}p{2.92cm}p{2.25cm}p{1.96cm}}
\hline
\multicolumn{1}{|p{3.08cm}}{\centering
\textbf{Author}} & 
\multicolumn{1}{|p{5.13cm}}{\centering
\textbf{Comedy}} & 
\multicolumn{1}{|p{2.92cm}}{\centering
\textbf{Date of composition}} & 
\multicolumn{1}{|p{2.25cm}}{\centering
\textbf{Source}} & 
\multicolumn{1}{|p{1.96cm}|}{\centering
\textbf{Number of verses}} \\ 
\hline
\multicolumn{1}{|p{3.08cm}}{\multirow{4}{*}{\parbox{3.08cm}{\centering
\textbf{Tirso de Molina}}}} & 
\multicolumn{1}{|p{5.13cm}}{\centering
\textit{El vergonzoso en palacio}} & 
\multicolumn{1}{|p{2.92cm}}{\centering
1610$-$1625} & 
\multicolumn{1}{|p{2.25cm}}{\centering
BVC} & 
\multicolumn{1}{|p{1.96cm}|}{\centering
3968} \\ 
\hhline{~----}
\multicolumn{1}{|p{3.08cm}}{} & 
\multicolumn{1}{|p{5.13cm}}{\centering
\textit{Don Gil de las calzas verdes}} & 
\multicolumn{1}{|p{2.92cm}}{\centering
1615} & 
\multicolumn{1}{|p{2.25cm}}{\centering
BVC} & 
\multicolumn{1}{|p{1.96cm}|}{\centering
3277} \\ 
\hhline{~----}
\multicolumn{1}{|p{3.08cm}}{} & 
\multicolumn{1}{|p{5.13cm}}{\centering
\textit{Palabras y plumas}} & 
\multicolumn{1}{|p{2.92cm}}{\centering
Previous to 1627} & 
\multicolumn{1}{|p{2.25cm}}{\centering
BVC} & 
\multicolumn{1}{|p{1.96cm}|}{\centering
3979} \\ 
\hhline{~----}
\multicolumn{1}{|p{3.08cm}}{} & 
\multicolumn{3}{|p{10.3cm}}{} & 
\multicolumn{1}{|p{1.96cm}|}{\centering
11224} \\ 
\hline
\multicolumn{1}{|p{3.08cm}}{\multirow{4}{*}{\parbox{3.08cm}{\centering
\textbf{Mira de Amescua}}}} & 
\multicolumn{1}{|p{5.13cm}}{\centering
\textit{Cautela contra cautela}} & 
\multicolumn{1}{|p{2.92cm}}{\centering
Previous to 1635} & 
\multicolumn{1}{|p{2.25cm}}{\centering
BVC} & 
\multicolumn{1}{|p{1.96cm}|}{\centering
2845} \\ 
\hhline{~----}
\multicolumn{1}{|p{3.08cm}}{} & 
\multicolumn{1}{|p{5.13cm}}{\centering
\textit{El esclavo del demonio}} & 
\multicolumn{1}{|p{2.92cm}}{\centering
1612} & 
\multicolumn{1}{|p{2.25cm}}{\centering
BVC} & 
\multicolumn{1}{|p{1.96cm}|}{\centering
3296} \\ 
\hhline{~----}
\multicolumn{1}{|p{3.08cm}}{} & 
\multicolumn{1}{|p{5.13cm}}{\centering
\textit{Hero y Leandro}} & 
\multicolumn{1}{|p{2.92cm}}{\centering
—} & 
\multicolumn{1}{|p{2.25cm}}{\centering
AHCT} & 
\multicolumn{1}{|p{1.96cm}|}{\centering
3310} \\ 
\hhline{~----}
\multicolumn{1}{|p{3.08cm}}{} & 
\multicolumn{3}{|p{10.3cm}}{} & 
\multicolumn{1}{|p{1.96cm}|}{\centering
9451} \\ 
\hline
\multicolumn{1}{|p{3.08cm}}{\multirow{4}{*}{\parbox{3.08cm}{\centering
\textbf{Andrés de Claramonte}}}} & 
\multicolumn{1}{|p{5.13cm}}{\centering
\textit{Deste agua no beberé}} & 
\multicolumn{1}{|p{2.92cm}}{\centering
Previous to 1617} & 
\multicolumn{1}{|p{2.25cm}}{\centering
BVC} & 
\multicolumn{1}{|p{1.96cm}|}{\centering
2743} \\ 
\hhline{~----}
\multicolumn{1}{|p{3.08cm}}{} & 
\multicolumn{1}{|p{5.13cm}}{\centering
\textit{El valiente negro en Flandes}} & 
\multicolumn{1}{|p{2.92cm}}{\centering
Previous to 1638} & 
\multicolumn{1}{|p{2.25cm}}{\centering
Bubok} & 
\multicolumn{1}{|p{1.96cm}|}{\centering
2973} \\ 
\hhline{~----}
\multicolumn{1}{|p{3.08cm}}{} & 
\multicolumn{1}{|p{5.13cm}}{\centering
\textit{Púsoseme el sol}} & 
\multicolumn{1}{|p{2.92cm}}{\centering
—} & 
\multicolumn{1}{|p{2.25cm}}{\centering
BVC} & 
\multicolumn{1}{|p{1.96cm}|}{\centering
3206} \\ 
\hhline{~----}
\multicolumn{1}{|p{3.08cm}}{} & 
\multicolumn{3}{|p{10.3cm}}{} & 
\multicolumn{1}{|p{1.96cm}|}{\centering
8922} \\ 
\hline
\multicolumn{1}{|p{3.08cm}}{\multirow{4}{*}{\parbox{3.08cm}{\centering
\textbf{Vélez de Guevara}}}} & 
\multicolumn{1}{|p{5.13cm}}{\centering
\textit{El diablo está en Cantillana}} & 
\multicolumn{1}{|p{2.92cm}}{\centering
1622} & 
\multicolumn{1}{|p{2.25cm}}{\centering
BVC} & 
\multicolumn{1}{|p{1.96cm}|}{\centering
2621} \\ 
\hhline{~----}
\multicolumn{1}{|p{3.08cm}}{} & 
\multicolumn{1}{|p{5.13cm}}{\centering
\textit{La serra de la Vera}} & 
\multicolumn{1}{|p{2.92cm}}{\centering
1613} & 
\multicolumn{1}{|p{2.25cm}}{\centering
BVC} & 
\multicolumn{1}{|p{1.96cm}|}{\centering
3306} \\ 
\hhline{~----}
\multicolumn{1}{|p{3.08cm}}{} & 
\multicolumn{1}{|p{5.13cm}}{\centering
\textit{También la afrenta es veneno }(acto primero)} & 
\multicolumn{1}{|p{2.92cm}}{\centering
—} & 
\multicolumn{1}{|p{2.25cm}}{\centering
BVC} & 
\multicolumn{1}{|p{1.96cm}|}{\centering
1196} \\ 
\hhline{~----}
\multicolumn{1}{|p{3.08cm}}{} & 
\multicolumn{3}{|p{10.3cm}}{} & 
\multicolumn{1}{|p{1.96cm}|}{\centering
7123} \\ 
\hline
\multicolumn{1}{|p{3.08cm}}{\multirow{5}{*}{\parbox{3.08cm}{\centering
\textbf{Unknown}}}} & 
\multicolumn{1}{|p{5.13cm}}{\centering
\textit{El burlador de Sevilla}} & 
\multicolumn{1}{|p{2.92cm}}{\centering
1612$-$1617} & 
\multicolumn{1}{|p{2.25cm}}{\centering
BVC} & 
\multicolumn{1}{|p{1.96cm}|}{\centering
2900} \\ 
\hhline{~----}
\multicolumn{1}{|p{3.08cm}}{} & 
\multicolumn{1}{|p{5.13cm}}{\centering
\textit{El condenado por desconfiado}} & 
\multicolumn{1}{|p{2.92cm}}{\centering
Previous to 1635} & 
\multicolumn{1}{|p{2.25cm}}{\centering
BVC} & 
\multicolumn{1}{|p{1.96cm}|}{\centering
2995} \\ 
\hhline{~----}
\multicolumn{1}{|p{3.08cm}}{} & 
\multicolumn{1}{|p{5.13cm}}{\centering
\textit{La mujer por fuerza}} & 
\multicolumn{1}{|p{2.92cm}}{\centering
Previous to 1635} & 
\multicolumn{1}{|p{2.25cm}}{\centering
Clásicos Hispánicos} & 
\multicolumn{1}{|p{1.96cm}|}{\centering
2886} \\ 
\hhline{~----}
\multicolumn{1}{|p{3.08cm}}{} & 
\multicolumn{1}{|p{5.13cm}}{\centering
\textit{La ninfa del cielo}} & 
\multicolumn{1}{|p{2.92cm}}{\centering
1610$-$1620} & 
\multicolumn{1}{|p{2.25cm}}{\centering
AHCT} & 
\multicolumn{1}{|p{1.96cm}|}{\centering
3505} \\ 
\hhline{~----}
\multicolumn{1}{|p{3.08cm}}{} & 
\multicolumn{1}{|p{5.13cm}}{\centering
\textit{Tan largo me lo fiáis}} & 
\multicolumn{1}{|p{2.92cm}}{\centering
1612$-$1617} & 
\multicolumn{1}{|p{2.25cm}}{\centering
BVC} & 
\multicolumn{1}{|p{1.96cm}|}{\centering
2760} \\ 
\hline
\end{tabular}
\end{adjustbox}\\
\caption{Plays in the corpus. Date of composition is specified as much as possible.}
\end{table}

Once the texts have been obtained and converted to plain text, they are pre$-$processed. The main modifications on the texts have been the following: codification in UTF$-$8, substitution of the line breaks of Windows for those of Linux (which do not imply carriage return), and the suppression of the numbering of the verses. The names of the play characters have not been excluded because onomastics has been one of the arguments that the researchers have put forward in their various proposals for attributions. Only after all this process is it possible to start experimenting with the texts.

\subsection{The experiments}
Two different types of experiments were carried out: clustering of plays and measuring distance between plays.

\subsubsection{Clustering analysis}
To configure the Stylo tool (version 0.6.9) properly we have carried out a set of preliminary tests with a reduced version of the corpus (eliminating works of unknown authorship). In this way we checked which values offered the most consistent results. We have concluded that the most appropriate value for the MFW parameter (most frequent words) was 250 for both maximum and minimum, in order to build one single dendogram. This is the essential point of the experiment, as variations on that figure may alter the results substantially. The features we decided to extract from the texts were unigrams of words, that is, tokens, and we did not consider necessary to perform any culling.  Then, we have applied the best configuration to the set of plays enumerated above in Table 3. The dendogram resulting from the experiment is depicted in Figure 1.

\begin{figure}
\centering
\includegraphics[width=17.0cm,height=17.0cm]{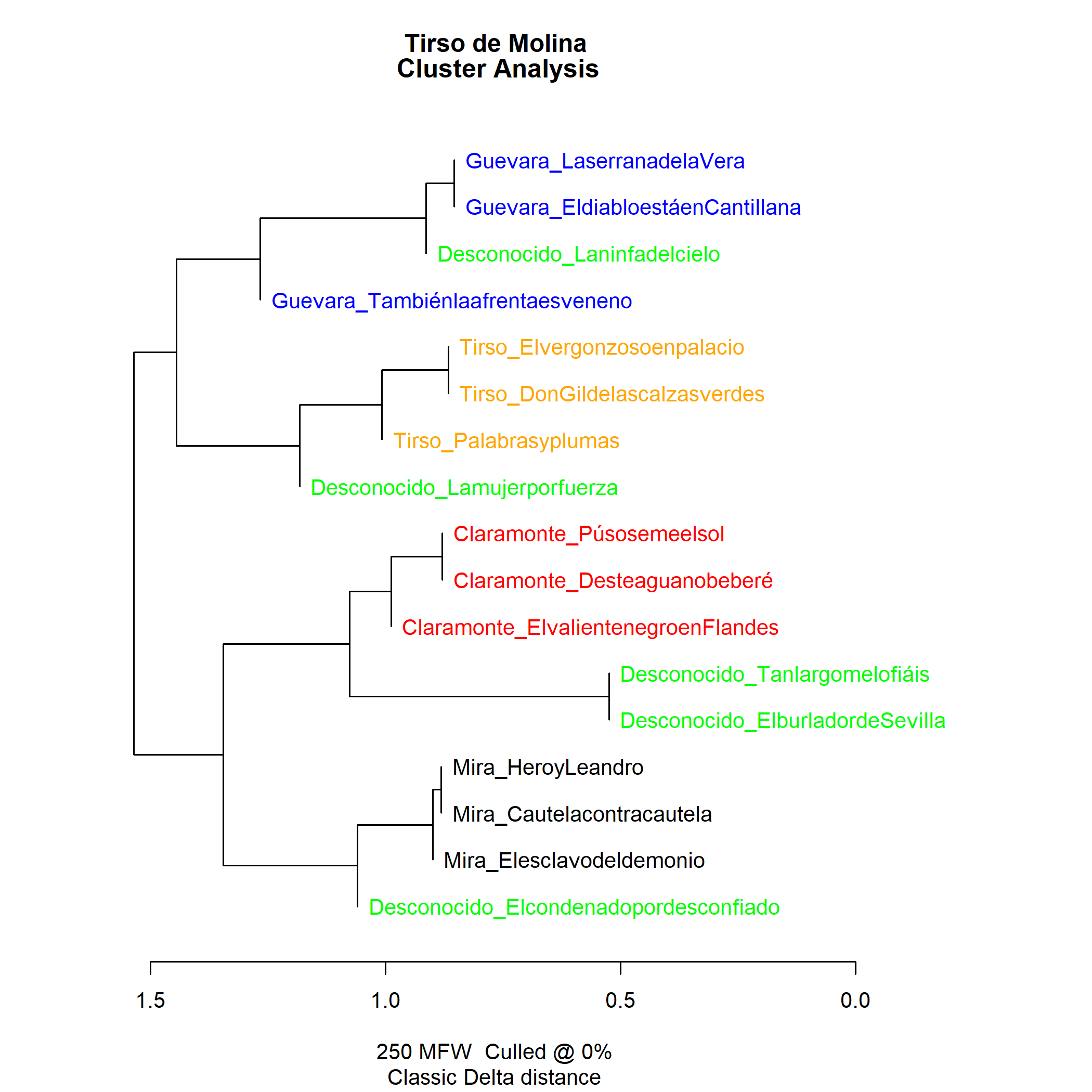}
\caption{Dendrogram of the clustering analysis using Delta Burrows distance.}
\end{figure}

As was to be expected, the smallest distance is that which separates \textit{El burlador de Sevilla} and \textit{Tan largo me lo fiáis}, since they are, in a high percentage, the same text. They are undoubtedly grouped together with Claramonte's production. Mira seems to be the author with the most homogenous style by virtue of the small distance between his three comedies of assured authorship, while Guevara presents the most unstable style. The pairs Guevara$-$Tirso and Claramonte$-$Mira are the ones that share the most stylistic similarities. As for the attributions of the plays of unknown authorship, the results obtained are categorical: \textit{El condenado por desconfiado}\ \ is associated with the plays of Mira, \textit{La mujer por fuerza} is agglutinated with the  plays of Tirso, and \textit{La ninfa del cielo} joins Guevara, the latter being the most categorical attribution of all.

As previously commented, the new Delta measure emended with Eder implementations (2016) yielded better results in some experiments. To contrast those obtained with the classic Burrows Delta, it is convenient to carry out another similar analysis but with that of Eder. This brings us to new results shown in the cluster analysis of Figure 2.

\begin{figure}
\centering
\includegraphics[width=17.0cm,height=17.0cm]{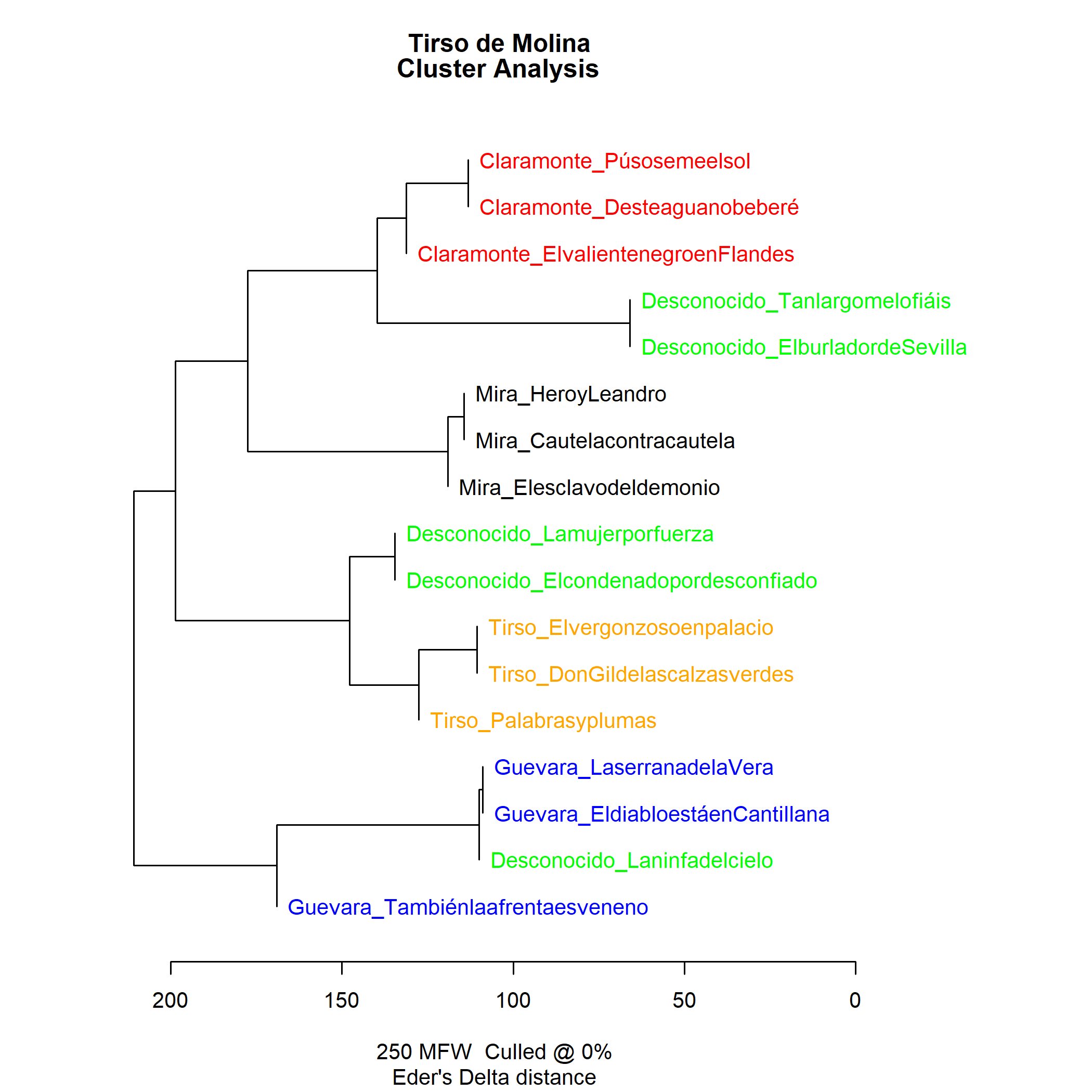}
\caption{Dendogram of the clustering analysis using Delta Eder distance.}
\end{figure}

The changes with respect to the previous dendogram are small but remarkable. The Claramonte$-$Mira pair still shares a high degree of similarity, but this is not the case with the pair formed by Tirso and Guevara, an author who now differs more from the other three. The main dissimilarity lies in the comedy \textit{El condenado por desconfiado}, which is very close in style to \textit{La mujer por fuerza} and is grouped like this with Tirso's works. The other attributions continue to maintain their rotundity, especially \textit{La ninfa del cielo}, which even reinforces it.

\subsubsection{Distance measures}

We have implemented an AAA free software tool to use the four distance measures defined in the previous section: Perplexity, Kullback$-$Leibler divergence, Ranking$-$based measure, and Distributional similarity. The results obtained, broken down by author and by measure, are shown in Table 4. 

\begin{table}[t]
\begin{adjustbox}{max width=\textwidth}
\begin{tabular}{p{2.67cm}p{2.76cm}p{2.76cm}p{2.76cm}p{2.76cm}p{2.84cm}}
\hline
\multicolumn{1}{|p{2.67cm}}{} & 
\multicolumn{1}{|p{2.76cm}}{\centering
\textbf{\textit{La ninfa del cielo}}} & 
\multicolumn{1}{|p{2.76cm}}{\centering
\textbf{\textit{El burlador de Sevilla}}} & 
\multicolumn{1}{|p{2.76cm}}{\centering
\textbf{\textit{Tan largo me lo fiáis}}} & 
\multicolumn{1}{|p{2.76cm}}{\centering
\textbf{\textit{La mujer por fuerza}}} & 
\multicolumn{1}{|p{2.84cm}|}{\centering
\textbf{\textit{El condenado por desconfiado}}} \\ 
\hline
\multicolumn{1}{|p{2.67cm}}{\multirow{4}{*}{\parbox{2.67cm}{\centering
\textbf{Perplexity}}}} & 
\multicolumn{1}{|p{2.76cm}}{\centering
0.0000 Mira} & 
\multicolumn{1}{|p{2.76cm}}{\centering
0.0000 Clar} & 
\multicolumn{1}{|p{2.76cm}}{\centering
0.0000 Clar} & 
\multicolumn{1}{|p{2.76cm}}{\centering
0.0000 Mira} & 
\multicolumn{1}{|p{2.84cm}|}{\centering
0.0000 Clar} \\ 
\hhline{~-----}
\multicolumn{1}{|p{2.67cm}}{} & 
\multicolumn{1}{|p{2.76cm}}{\centering
0.0483 Tirso} & 
\multicolumn{1}{|p{2.76cm}}{\centering
0.1525 Tirso} & 
\multicolumn{1}{|p{2.76cm}}{\centering
0.5612 Tirso} & 
\multicolumn{1}{|p{2.76cm}}{\centering
0.1739 Tirso} & 
\multicolumn{1}{|p{2.84cm}|}{\centering
0.2502 Mira} \\ 
\hhline{~-----}
\multicolumn{1}{|p{2.67cm}}{} & 
\multicolumn{1}{|p{2.76cm}}{\centering
0.4193 Clar} & 
\multicolumn{1}{|p{2.76cm}}{\centering
0.6602 Mira} & 
\multicolumn{1}{|p{2.76cm}}{\centering
0.7078 Guev} & 
\multicolumn{1}{|p{2.76cm}}{\centering
0.7770 Clar} & 
\multicolumn{1}{|p{2.84cm}|}{\centering
0.6701 Tirso} \\ 
\hhline{~-----}
\multicolumn{1}{|p{2.67cm}}{} & 
\multicolumn{1}{|p{2.76cm}}{\centering
1.0000 Guev} & 
\multicolumn{1}{|p{2.76cm}}{\centering
1.0000 Guev} & 
\multicolumn{1}{|p{2.76cm}}{\centering
1.0000 Mira} & 
\multicolumn{1}{|p{2.76cm}}{\centering
1.0000 Guev} & 
\multicolumn{1}{|p{2.84cm}|}{\centering
1.0000 Guev} \\ 
\hline
\multicolumn{1}{|p{2.67cm}}{\multirow{4}{*}{\parbox{2.67cm}{\centering
\textbf{Kullback Leibler}}}} & 
\multicolumn{1}{|p{2.76cm}}{\centering
0.0000 Mira} & 
\multicolumn{1}{|p{2.76cm}}{\centering
0.0000 Clar} & 
\multicolumn{1}{|p{2.76cm}}{\centering
0.0000 Clar} & 
\multicolumn{1}{|p{2.76cm}}{\centering
0.0000 Tirso} & 
\multicolumn{1}{|p{2.84cm}|}{\centering
0.0000 Mira} \\ 
\hhline{~-----}
\multicolumn{1}{|p{2.67cm}}{} & 
\multicolumn{1}{|p{2.76cm}}{\centering
0.5770 Clar} & 
\multicolumn{1}{|p{2.76cm}}{\centering
0.5322 Guev} & 
\multicolumn{1}{|p{2.76cm}}{\centering
0.1294 Tirso} & 
\multicolumn{1}{|p{2.76cm}}{\centering
0.1626 Mira} & 
\multicolumn{1}{|p{2.84cm}|}{\centering
0.5719 Clar} \\ 
\hhline{~-----}
\multicolumn{1}{|p{2.67cm}}{} & 
\multicolumn{1}{|p{2.76cm}}{\centering
0.6636 Guev} & 
\multicolumn{1}{|p{2.76cm}}{\centering
0.6288 Tirso} & 
\multicolumn{1}{|p{2.76cm}}{\centering
0.8314 Guev} & 
\multicolumn{1}{|p{2.76cm}}{\centering
0.6457 Guev} & 
\multicolumn{1}{|p{2.84cm}|}{\centering
0.8423 Guev} \\ 
\hhline{~-----}
\multicolumn{1}{|p{2.67cm}}{} & 
\multicolumn{1}{|p{2.76cm}}{\centering
10000 Tirso} & 
\multicolumn{1}{|p{2.76cm}}{\centering
1.0000 Mira} & 
\multicolumn{1}{|p{2.76cm}}{\centering
1.0000 Mira} & 
\multicolumn{1}{|p{2.76cm}}{\centering
1.0000 Clar} & 
\multicolumn{1}{|p{2.84cm}|}{\centering
1.0000 Tirso} \\ 
\hline
\multicolumn{1}{|p{2.67cm}}{\multirow{4}{*}{\parbox{2.67cm}{\centering
\textbf{Ranking}}}} & 
\multicolumn{1}{|p{2.76cm}}{\centering
0.0000 Mira} & 
\multicolumn{1}{|p{2.76cm}}{\centering
0.0000 Clar} & 
\multicolumn{1}{|p{2.76cm}}{\centering
0.0000 Clar} & 
\multicolumn{1}{|p{2.76cm}}{\centering
0.0000 Tirso} & 
\multicolumn{1}{|p{2.84cm}|}{\centering
0.0000 Clar} \\ 
\hhline{~-----}
\multicolumn{1}{|p{2.67cm}}{} & 
\multicolumn{1}{|p{2.76cm}}{\centering
0.4000 Guev} & 
\multicolumn{1}{|p{2.76cm}}{\centering
0.7895 Guev} & 
\multicolumn{1}{|p{2.76cm}}{\centering
0.0238 Mira} & 
\multicolumn{1}{|p{2.76cm}}{\centering
0.2149 Clar} & 
\multicolumn{1}{|p{2.84cm}|}{\centering
0.2410 Mira} \\ 
\hhline{~-----}
\multicolumn{1}{|p{2.67cm}}{} & 
\multicolumn{1}{|p{2.76cm}}{\centering
0.8828 Clar} & 
\multicolumn{1}{|p{2.76cm}}{\centering
0.8684 Tirso} & 
\multicolumn{1}{|p{2.76cm}}{\centering
0.3333 Guev} & 
\multicolumn{1}{|p{2.76cm}}{\centering
0.6033 Mira} & 
\multicolumn{1}{|p{2.84cm}|}{\centering
0.2530 Tirso} \\ 
\hhline{~-----}
\multicolumn{1}{|p{2.67cm}}{} & 
\multicolumn{1}{|p{2.76cm}}{\centering
1.0000 Tirso} & 
\multicolumn{1}{|p{2.76cm}}{\centering
1.0000 Mira} & 
\multicolumn{1}{|p{2.76cm}}{\centering
1.0000 Tirso} & 
\multicolumn{1}{|p{2.76cm}}{\centering
1.0000 Guev} & 
\multicolumn{1}{|p{2.84cm}|}{\centering
1.0000 Guev} \\ 
\hline
\multicolumn{1}{|p{2.67cm}}{\multirow{4}{*}{\parbox{2.67cm}{\centering
\textbf{Distributional}}}} & 
\multicolumn{1}{|p{2.76cm}}{\centering
0.0000 Mira} & 
\multicolumn{1}{|p{2.76cm}}{\centering
0.0000 Clar} & 
\multicolumn{1}{|p{2.76cm}}{\centering
0.0000 Clar} & 
\multicolumn{1}{|p{2.76cm}}{\centering
0.0000 Tirso} & 
\multicolumn{1}{|p{2.84cm}|}{\centering
0.0000 Mira} \\ 
\hhline{~-----}
\multicolumn{1}{|p{2.67cm}}{} & 
\multicolumn{1}{|p{2.76cm}}{\centering
0.3204 Guev} & 
\multicolumn{1}{|p{2.76cm}}{\centering
0.0805 Mira} & 
\multicolumn{1}{|p{2.76cm}}{\centering
0.4545 Mira} & 
\multicolumn{1}{|p{2.76cm}}{\centering
0.0058 Mira} & 
\multicolumn{1}{|p{2.84cm}|}{\centering
0.1944 Clar} \\ 
\hhline{~-----}
\multicolumn{1}{|p{2.67cm}}{} & 
\multicolumn{1}{|p{2.76cm}}{\centering
0.4757 Tirso} & 
\multicolumn{1}{|p{2.76cm}}{\centering
0.6356 Tirso} & 
\multicolumn{1}{|p{2.76cm}}{\centering
0.6098 Tirso} & 
\multicolumn{1}{|p{2.76cm}}{\centering
0.1930 Clar} & 
\multicolumn{1}{|p{2.84cm}|}{\centering
0.4259 Tirso} \\ 
\hhline{~-----}
\multicolumn{1}{|p{2.67cm}}{} & 
\multicolumn{1}{|p{2.76cm}}{\centering
1.0000 Clar} & 
\multicolumn{1}{|p{2.76cm}}{\centering
1.0000 Guev} & 
\multicolumn{1}{|p{2.76cm}}{\centering
1.0000 Guev} & 
\multicolumn{1}{|p{2.76cm}}{\centering
1.0000 Guev} & 
\multicolumn{1}{|p{2.84cm}|}{\centering
1.0000 Guev} \\ 
\hline
\end{tabular}
\end{adjustbox}\\
\caption{Normalized results obtained comparing doubtful authorship plays with authors under discussion. Claramonte and Guevara’s names are shortend in order to unify the cells' size.}
\end{table}

Perplexity discards the authorship of Tirso for all the works under study; however it shows him as a second option, very close to the first one, for most of them. The position of Tirso in the other measures tends to be less favorable. As the most distant author, Perplexity points in almost all cases to Guevara, including curiously that of \textit{La ninfa del cielo}, whose attribution to Guevara was to be expected.  

Kullback$-$Leibler agrees with Stylo except for \textit{La ninfa del cielo}, which is attributed to Mira, although this measure places Guevara as the second most plausible author of this play.

Ranking$-$based measure proposes Claramonte as the most possible author of \textit{El condenado}, although the distance with Mira and Tirso is small, and the distance between these two authors is still smaller, in contrast to the notable difference that Perplexity makes between them for the aforementioned comedy.  However, the main dissimilarity between Ranking and Perplexity lies in \textit{La mujer por fuerza}, which Perplexity, as opposed to the other measures, attributes to Mira. 

Distributional similarity presents as first option the same authors as Kullback-Leibler, but for the other options it tends to give more chances to Mira. 

Between \textit{El burlador} and \textit{Tan Largo}, the four measures show more clarity with \textit{El burlador}, but they are quite clear and unanimous in the attribution to Claramonte of both versions. The consensus among the four measures is remarkable although the values of Perplexity diverge slightly from the other measures. The reason for this slight disparity is to be found in the types of n$-$grams that the strategies extract from the texts: while all the others use word unigrams, Perplexity uses 7$-$grams of characters. Table 5 shows the normalized average scores of the four measures.

\begin{table}[h]
\begin{adjustbox}{max width=\textwidth}
\begin{tabular}{p{2.9cm}p{2.9cm}p{2.9cm}p{2.9cm}p{2.99cm}}
\hline
\multicolumn{1}{|p{2.9cm}}{\centering
\textbf{\textit{La ninfa del cielo}}} & 
\multicolumn{1}{|p{2.9cm}}{\centering
\textbf{\textit{El burlador de Sevilla}}} & 
\multicolumn{1}{|p{2.9cm}}{\centering
\textbf{\textit{Tan largo me lo fiáis}}} & 
\multicolumn{1}{|p{2.9cm}}{\centering
\textbf{\textit{La mujer por fuerza}}} & 
\multicolumn{1}{|p{2.99cm}|}{\centering
\textbf{\textit{El condenado por desconfiado}}} \\ 
\hline
\multicolumn{1}{|p{2.9cm}}{\centering
0.000 Mira} & 
\multicolumn{1}{|p{2.9cm}}{\centering
0.000 Cla} & 
\multicolumn{1}{|p{2.9cm}}{\centering
0.000 Clar} & 
\multicolumn{1}{|p{2.9cm}}{\centering
0.043 Tirso} & 
\multicolumn{1}{|p{2.99cm}|}{\centering
0.123 Mira} \\ 
\hline
\multicolumn{1}{|p{2.9cm}}{\centering
0.596 Guev} & 
\multicolumn{1}{|p{2.9cm}}{\centering
0.571 Tirso} & 
\multicolumn{1}{|p{2.9cm}}{\centering
0.575 Tirso} & 
\multicolumn{1}{|p{2.9cm}}{\centering
0.193 Mira} & 
\multicolumn{1}{|p{2.99cm}|}{\centering
0.192 Clar} \\ 
\hline
\multicolumn{1}{|p{2.9cm}}{\centering
0.631 Tirso} & 
\multicolumn{1}{|p{2.9cm}}{\centering
0.685 Mira} & 
\multicolumn{1}{|p{2.9cm}}{\centering
0.620 Mira} & 
\multicolumn{1}{|p{2.9cm}}{\centering
0.546 Clar} & 
\multicolumn{1}{|p{2.99cm}|}{\centering
0.587 Tirso} \\ 
\hline
\multicolumn{1}{|p{2.9cm}}{\centering
0.720 Clar} & 
\multicolumn{1}{|p{2.9cm}}{\centering
0.830 Guev} & 
\multicolumn{1}{|p{2.9cm}}{\centering
0.718 Guev} & 
\multicolumn{1}{|p{2.9cm}}{\centering
0.911 Guev} & 
\multicolumn{1}{|p{2.99cm}|}{\centering
0.961 Guev} \\ 
\hline
\end{tabular}
\end{adjustbox}\\
\caption{Mean of the results obtained comparing doubtful authorship plays with authors under discussion.}
\end{table}

\textit{La ninfa del cielo}, whose authorship for Guevara was established as quite sure by experts, does not seem to come so close to the style of this author, but rather presents more confluences with Mira de Amescua. The results for \textit{El burlador de Sevilla} and \textit{Tan largo me lo fiáis} are sufficiently categorical and similar across all measures to reject Tirso de Molina's supposed authorship. All the measures aim to support the position of critics who defend the authorship of Claramonte. \textit{La mujer por fuerza} and \textit{El condenado por desconfiado} are the plays that leave more room for doubt and alternative hypotheses, since the four distances differ in small values from each other. In fact, the distance between the first and second authors is less than or equal to 0.15 in all cases. Mira and Tirso are the most likely authors of these two comedies.

\subsection{Discussion}
It is now up to us to carry out a joint comparison of the various results obtained by all the strategies. This will allow us to draw conclusions in order to confirm or not the authorship of the texts. Authorships proposed by the different methods, including traditional philological studies, are illustrated in Table 6.

\begin{table}[h]
\begin{adjustbox}{max width=\textwidth}
\begin{tabular}{p{4.01cm}p{5.77cm}p{2.36cm}p{3.33cm}}
\hline
\multicolumn{1}{|p{4.01cm}}{} & 
\multicolumn{1}{|p{5.77cm}}{\centering
\textbf{Philological studies}} & 
\multicolumn{1}{|p{2.36cm}}{\centering
\textbf{Clustering analysis}} & 
\multicolumn{1}{|p{3.33cm}|}{\centering
\textbf{Distance measures}} \\ 
\hline
\multicolumn{1}{|p{4.01cm}}{\centering
\textbf{\textit{La ninfa del cielo}}} & 
\multicolumn{1}{|p{5.77cm}}{\centering
Tirso / Guevara} & 
\multicolumn{1}{|p{2.36cm}}{\centering
Guevara} & 
\multicolumn{1}{|p{3.33cm}|}{\centering
Mira} \\ 
\hline
\multicolumn{1}{|p{4.01cm}}{\centering
\textbf{\textit{El burlador de Sevilla}}} & 
\multicolumn{1}{|p{5.77cm}}{\centering
Tirso / Claramonte} & 
\multicolumn{1}{|p{2.36cm}}{\centering
Claramonte} & 
\multicolumn{1}{|p{3.33cm}|}{\centering
Claramonte} \\ 
\hline
\multicolumn{1}{|p{4.01cm}}{\centering
\textbf{\textit{Tan largo me lo fiáis}}} & 
\multicolumn{1}{|p{5.77cm}}{\centering
Tirso / Claramonte} & 
\multicolumn{1}{|p{2.36cm}}{\centering
Claramonte} & 
\multicolumn{1}{|p{3.33cm}|}{\centering
Claramonte} \\ 
\hline
\multicolumn{1}{|p{4.01cm}}{\centering
\textbf{\textit{La mujer por fuerza}}} & 
\multicolumn{1}{|p{5.77cm}}{\centering
Tirso / Otro} & 
\multicolumn{1}{|p{2.36cm}}{\centering
Tirso} & 
\multicolumn{1}{|p{3.33cm}|}{\centering
Tirso} \\ 
\hline
\multicolumn{1}{|p{4.01cm}}{\centering
\textbf{\textit{El condenado por desconfiado}}} & 
\multicolumn{1}{|p{5.77cm}}{\centering
Tirso / Claramonte / Guevara / Mira / Colaboration} & 
\multicolumn{1}{|p{2.36cm}}{\centering
Mira} & 
\multicolumn{1}{|p{3.33cm}|}{\centering
Mira} \\ 
\hline
\end{tabular}
\end{adjustbox}\\
\caption{Authorship attributions of the plays under discussion proposed by the strategies.}
\end{table}

\textit{La ninfa del cielo}, whose restitution to Guevara seemed definitive, does not obtain the expected results with the distance measures, although Stylo's dendograms do group it with the rest of Guevara's works, being also the most categorical attribution of all. Although one cannot exclude the figure of Mira de Amescua, often neglected in the critical debates of the Golden Age theatre, it does not seem very likely to be behind this comedy. Hence, the arguments of the critics in favor of Guevara are strengthened by our study.

\textit{El burlador} and \textit{Tan largo}, the most important works that we are analyzing, obtain very enlightening results. The hypothesis insistently defended by Rodríguez López$-$Vázquez for so many years is very likely to be true. All the measures point to the same author: Andrés de Claramonte, a figure ignored and denigrated by the critics who, in view of these results, deserves at least a revaluation of his work and the reconsideration of his position in the canon of Spanish literature. It should not be affirmed, from this study, that Claramonte is indeed the author of the first don Juan, but it can be assured that Tirso de Molina's most famous work was not written by Tirso de Molina, since the other authors inserted in the corpus (Guevara and Mira) have a style as close to the famous comedy as that of Tirso.

\textit{La mujer por fuerza} and \textit{El condenado por desconfiado} must be considered together. As previously explained (see Section 4.1), one of them is the fourth work of Tirso in his second block of comedies, so they cannot both belong to Tirso. In view of the fact that no other author has been proposed for \textit{La mujer por fuerza} and the results it has obtained, it seems most likely that this is the fourth work that does belong to Tirso. So many alternatives have been proposed for \textit{El condenado} that it is risky to bet on one. Although this study situates Mira as the most probable candidate and rules out, because of the previous attribution, Tirso, it is also necessary to consider as feasible the hypothesis of collaboration, which cannot be verified with the strategies employed in our study.

In short, Tirso has been attributed a considerable number of works based on conjectures and critical arguments lacking in solidity and documentary proof. So far, a high percentage of the production plays traditionally assigned to Tirso does not actually belong to this author. In the 17\textsuperscript{th} century comedies were published under the name of famous authors in order to increase sales; it seems that subsequent literary critics have been carried away by this personalist tendency by favoring attributions to renowned authors. The curious thing about Tirso de Molina's case is that these false attributions have been elaborated one on top of the other, in such a way that questioning one implies questioning them all. It is important to point out that the controversial plays are precisely those on which the fame of Tirso among the public and his excellent critical appraisal are based. So perhaps the place occupied by this playwright in the history of Spanish literature should begin to be reconsidered. It is more urgent, however, to draw attention to the traditional studies of authorship attribution, which on many occasions have not respected the basic principles of scientific rigor that should govern any kind of humanistic research.

\section{Conclusions and future work}\label{sec:conclusions}
After combining traditional and non$-$traditional studies for a correct interpretation of the results, the rotundity of these forces us to position ourselves in favor of those theories that propose less famous authors who have been relegated to a second place in the panorama of Golden Age theatre.  Among these less famous authors, Mira de Amescua, Vélez de Guevara and, especially, Andrés de Claramonte should be pointed out. The conclusive proof that Claramonte was the author of the first don Juan seems to be closer than ever. An act of justice would be to vindicate his work, starting by editing it. Yet, in spite of the evidence shown by our study, in order to be even more certain of the results, it will be still necessary to approach this problem with new studies that use more texts, more authors and more advanced NLP strategies such as those based on distributional semantics and contextualized word embeddings (Gamallo 2019).

In future work, the comedy \textit{El condenado por desconfiado }would deserve a separate study, as its authorship hypotheses are too varied and confusing to fit into our current work. For this purpose, it will be necessary to employ tools that study pieces of texts separately so as to determine if it is a work composed in collaboration or if one of the proposed authors is indeed the authentic one. The development and improvement of the AAA tools is, in fact, another step to follow in order to continue deepening the AAA studies. This is the most promising point of our work, since we explored computational strategies that are useful and efficient in this task, even though their original functionality was not the quantification of style for authorship attribution. We think that, in the future, it will be possible to find other techniques and strategies that fit well within the AAA studies. 

In any case, the most urgent initiative that should be encouraged is the edition and digitization of Golden Age plays, so that it is possible to replicate our experiment on the work of Tirso on a large scale.

\begin{enumerate}
	\item[]{\large  \bf{References}}

\end{enumerate}

Almeida, D. C. (2014). Atribuição de autoria com propósitos forenses: panorama e proposta de análise. \textit{ReVEL: Revista Virtual de Estudos da Linguagem}, \textit{12(23),}\ \ 148$-$186. Retrieved from \url{http://www.revel.inf.br/files/539b2f0878d56cb6604363c111dfe116.pdf}

Argamon, S. $\&$ Juola, P. (2011). \textit{Overview of the International Authorship Identification Competition at PAN$-$2011}. CLEF 2011 Labs and Workshop: Amsterdam.

Azfali, M. $\&$ Kumar, S. (2017). Comparative Analysis of Various Similarity Measures for Finding Similarity of Two Documents. \textit{International Journal of Database Theory and Application, 10(2),} 23$-$30.

Blasco, J. (2016). Avellaneda desde la estilometría. In P. Ruiz (ed), \textit{Cervantes: los viajes y los días }(pp. 97$-$116), Madrid: Sial Ediciones. 

Burrows, J. (2002). $``$Delta$"$: a measure of stylistic difference and a guide to likely authorship. \textit{Literary and Linguistic Computing}, \textit{17(3),} 267$-$287. Retrieved from \url{https://academic.oup.com/dsh/article/17/3/267/929277}

Calvo Tello, J. (2016). Entendiendo Delta desde las Humanidades. \textit{Caracteres, Estudios culturales y críticos de la esfera digital, 5(1),} 140$-$176. 

Cavnar, W. B. $\&$ Trenkle, J. M. (1994). N$-$gram$-$based text categorization. \textit{Proceedings of the Third Symposium on Document Analysis and Information Retrieval}, Las Vegas: University of Nevada.

Cuéllar González, Á. $\&$ García$-$Luengos, G. V. (2017). EstilometríaTSO: Estilometría aplicada al teatro del Siglo de Oro \url{http://estilometriatso.com/}

Eder, M., Rybicki, J. $\&$ Kestemont, M. (2016). Stylometry with R: a package for computational text analysis. \textit{R Journal, 8(1),} 107-121. Retrieved from \url{https://journal.r-project.org/archive/2016/RJ-2016-007/index.html}

Eisen, M., Ribeiro, A., Segarra, S. $\&$ Egan, G. (2018). Stylometric analysis of Early Modern period English plays. \textit{Digital Scholarship in the Humanities, 33(3),} 500$-$528. doi:10.1093/llc/fqx059

Gamallo, P., Garcia, M., Sotelo, S. $\&$ Pichel, J. R. (2014). Comparing Ranking-based and Naive Bayes Approaches to Language Detection on Tweets. \textit{Proceedings of XXX Congreso de la Sociedad Española de Procesamiento de lenguaje natural}, Girona: SEPLN.

Gamallo, P., Pichel, J. R. $\&$ Alegria, I. (2017). From language identification to language distance. \textit{Physica A}, 484, 162$-$172. 

Gamallo, P., Sotelo, S., Pichel, J. R. $\&$ Artexte, M. (2019). Contextualized Translations of Phrasal Verbs with Distributional Compositional Semantics and Monolingual Corpora. \textit{Journal of Computational Linguistics, 45(3)}, 395$-$421.

García Gómez, Á. M. (2005). Aporte documental al debate acerca de la prioridad entre \textit{El burlador de Sevilla }y \textit{Tan largo me lo fiáis}: el cartapacio de comedias de Jerónimo Sánchez. In A. J. Close and S. M. Fernández Valez (eds), \textit{Edad de oro cantabrigense: actas del VII Congreso de la Asociación Internacional de Hispanistas del Siglo de Oro, Madrid: Asociación Internacional del Siglo de Oro} (pp. 281$-$286).

García$-$Reidy, A. (2019). Deconstructing the Authorship of \textit{Siempre ayuda la verdad}: A Play by Lope de Vega?. \textit{Neophilologus}, 1$-$18. Retrieved from \url{https://link.springer.com/content/pdf/10.1007\%2Fs11061-019-09607-8.pdf}

Grieve, J. (2005). \textit{Quantitative Authorship Attribution: a History and an Evaluation of Techniques. }Burnaby: Simon Fraser University.

Ilsemann, H. (2018). Stylometry approaching Parnassus. \textit{Digital Scholarship in the Humanities, 33(3)}, 548$-$556.{\footnotesize  }doi:10.1093/llc/fqx058

Ilsemann, H. (2019). Forensic sylometry. \textit{Digital Scholarship in the Humanities, 34(2),} 335$-$349.{\footnotesize  }doi:10.1093/llc/fqy023

Iriarte, Á., Gamallo, P. $\&$ Simões, A. (2018). Estratégias Lexicométricas para Detetar Especificidades Textuais. \textit{Linguamática, 10(1),} 19$-$26. Retrieved from \url{https://linguamatica.com/index.php/linguamatica/article/view/263}

La Rosa, J. $\&$ Suárez, J. L. (2016). The Life of \textit{Lazarillo de Tormes} and of His Machine Learning Adversities: Non$-$traditional authorship attribution techniques in the context of the \textit{Lazarillo}. \textit{Lemir: Revista de Literatura Española Medieval y del Renacimiento, 20}, 373$-$438. 

Oteiza, B. (2000). ¿Conocemos los textos verdaderos de Tirso de Molina?. In I. Arellano and B. Oteiza (eds), \textit{Varia lección de Tirso de Molina (Actas del VIII Seminario del Centro para la Edición de Clásicos Españoles)} (pp. 99$-$128). Pamplona: Instituto de Estudios Tirsianos.

Pichel, J., Gamallo, P. $\&$ Alegría, I. (2019). Measuring diachronic language distance using perplexity: Application to English, Portuguese, and Spanish. \textit{Natural Language Engineering,} 1$-$22. doi:10.1017/S1351324919000378

Potha, N. $\&$ Stamatos, E. (2014). A Profile$-$Based Method for Authorship Verification. In A. Likas, K. Blekas, and D. Kalles (eds), \textit{Artificial Intelligence: Methods and Applications (SETN 2014, Lecture Notes in Computer Science) }(pp. 313$-$326). Retrieved from \url{https://link.springer.com/chapter/10.1007\%2F978-3-319-07064-3_25}

Rodríguez López$-$Vázquez, A. (1990). El estado de la cuestión en torno a Claramonte y \textit{El burlador de Sevilla}. \textit{Murgetana, 82}, 5$-$22.

Rodríguez López$-$Vázquez, A. (2010). \textit{La mujer por fuerza, El condenado por desconfiado }y \textit{El burlador de Sevilla}, tres comedias atribuidas a Tirso de Molina. \textit{Castilla: Estudios de Literatura, 1}, 131$-$153.

Rudman, J. (1998). The State of Authorship Attribution Studies: Some Problems and Solutions. \textit{Computers and the Humanities, 31(4)}, 351$-$365.

Rudman, J. (2016). Non$-$Traditional Authorship Attribution Studies of William Shakespeare’s Canon: Some Caveats. \textit{Journal of Early Modern Studies, 5,} 307$-$328. Retrieved from \url{http://www.fupress.net/index.php/bsfm-jems/article/view/18094/16848}

Stamatos, E. (2009). A Survey of Modern Authorship Attribution Methods. \textit{Journal of the American Society for Information Science and Technology, 60(3), }538$-$556. Retrieved from \url{http://citeseerx.ist.psu.edu/viewdoc/download?doi=10.1.1.440.1634&rep=rep1&type=pdf}

Vázquez, L. (1995). \textit{El burlador de Sevilla}: claramente de Tirso y no de Claramonte (breve anotación crítica). \textit{Bulletin of the comediantes, 47(2),} 183$-$190.

Vickers, B. (2011). Review: Shakespeare and Authorship Studies in the Twenty-First Century. \textit{Shakespeare Quarterly, 62(1),} 106$-$142.

\end{document}